\documentclass[letterpaper]{article} 
\usepackage{aaai2026}  
\usepackage{times}  
\usepackage{helvet}  
\usepackage{courier}  
\usepackage[hyphens]{url}  
\usepackage{graphicx} 
\usepackage{natbib}  
\usepackage{caption} 
\usepackage{algorithm}
\usepackage{algorithmic}

\usepackage{soul}
\usepackage{url}
\usepackage[utf8]{inputenc}
\usepackage{amsmath}
\usepackage{amsthm}
\usepackage{booktabs}
\usepackage{subcaption}
\usepackage{amssymb}
\usepackage{multirow}
\usepackage{xcolor}
\usepackage{makecell}
\usepackage{adjustbox}
\usepackage{threeparttable}

\definecolor{c1}{RGB}{68, 114, 196}
\definecolor{c2}{RGB}{0, 0, 0}
\definecolor{lightblue}{rgb}{0.7,0.9,1}

\newcommand{\first}[1]{\textbf{{\color{c1}#1}}}
\newcommand{\second}[1]{{\textbf{\color{c2}#1}}}

\usepackage{newfloat}
\usepackage{listings}
\DeclareCaptionStyle{ruled}{labelfont=normalfont,labelsep=colon,strut=off} 
\floatstyle{ruled}
\newfloat{listing}{tb}{lst}{}
\floatname{listing}{Listing}
\title{GT-SNT: A Linear-Time Transformer for Large-Scale Graphs via Spiking Node Tokenization}
\author{
    Huizhe Zhang\textsuperscript{\rm 1}, Jintang Li\textsuperscript{\rm 2}, Yuchang Zhu\textsuperscript{\rm 1}, Huazhen Zhong\textsuperscript{\rm 1}, Liang Chen\textsuperscript{\rm 1} \thanks{Corresponding author.}
}
\affiliations{
    \textsuperscript{\rm 1}Sun Yat-sen University\\
    \textsuperscript{\rm 2}Xiamen University\\

    zhanghzh33@mail2.sysu.edu.cn, edisonleejt@gmail.com, zhuych27@mail2.sysu.edu.cn, \\ zhonghzh9@mail2.sysu.edu.cn, chenliang6@mail.sysu.edu.cn
}

\usepackage{bibentry}

\begin{document}

\maketitle

\begin{abstract}
Graph Transformers (GTs), which integrate message passing and self-attention mechanisms simultaneously, have achieved promising empirical results in graph prediction tasks. However, the design of scalable and topology-aware node tokenization has lagged behind other modalities. This gap becomes critical as the quadratic complexity of full attention renders them impractical on large-scale graphs.
Recently, Spiking Neural Networks (SNNs), as brain-inspired models, provided an energy-saving scheme to convert input intensity into discrete spike-based representations through event-driven spiking neurons.
Inspired by these characteristics, we propose a linear-time \textbf{G}raph \textbf{T}ransformer with \textbf{S}piking \textbf{N}ode \textbf{T}okenization (GT-SNT) for node classification. By integrating multi-step feature propagation with SNNs, spiking node tokenization generates compact, locality-aware spike count embeddings as node tokens to avoid predefined codebooks and their utilization issues. The codebook guided self-attention leverages these tokens to perform node-to-token attention for linear-time global context aggregation.
In experiments, we compare GT-SNT with other state-of-the-art baselines on node classification datasets ranging from small to large. Experimental results show that GT-SNT achieves comparable performances on most datasets and reaches up to \textbf{130×} faster inference speed compared to other GTs. 
\end{abstract}

\begin{links}
    \link{Code}{https://github.com/Zhhuizhe/GT-SNT}
\end{links}

\section{Introduction}

Graph Transformers (GTs), as emerging graph representation learning paradigms, are proposed for alleviating inherent drawbacks of message passing neural networks like over-smoothing, over-squashing and local structure biases \cite{oono2019graph, topping2021understanding}. 
Benefiting from the multi-head attention (MHA) modules, vanilla Transformers adaptively learn the global dependencies of input sequences \cite{vaswani2017attention}. It also provides a solution for learning new topology among nodes while performing message aggregation on the graph data.  
Experiments demonstrate the immense potential of Transformers in handling global or long-range interactions \cite{rampavsek2022recipe, bo2023specformer}. However, there is one critical drawback that Transformers with $O(N^2)$ computation complexity are prohibitive for large-scale graphs.
Although empirical experiments show that designing linear-time attention in GTs can significantly reduce redundant computational cost \cite{wu2022nodeformer, wu2024sgformer}, such approaches introduce potential issues related to over-globalization \cite{xing2024less}. As an orthogonal technique, encapsulating structural or semantic information into node tokens provides a promising direction that enables GTs to efficiently capture salient graph properties via complementary local and global tokens \cite{wang2025gqt, luo2024nodeid}.

Recently, with the development of neuromorphic computing, Spiking Neural Networks (SNNs) offer a promising path toward energy-saving neural networks that achieve competitive performances using as few computations as possible. The defining feature of SNNs is the brain-like spiking mechanism which converts real-value signals into single-bit, sparse spiking signals based on event-driven biological neurons. The single-bit nature enables us to adopt more addition operations rather than expensive multiply-and-accumulate operations on the spiking outputs. Besides, the sparsity means spikes are cheap to store \cite{eshraghian2023training}. These delightful characteristics have prompted some studies constructing lightweight Spiking Graph Neural Networks (SGNNs) to explore spike-based representations on the graph data \cite{zhu2022spiking,li2023scaling, li2023graph}.
Despite tremendous progress in replacing artificial neurons with spiking counterparts, the broader use of SNNs for graph representation learning remains underexplored. Existing SGNNs tend to focus on their low-power advantages, while overlooking the expressive power of their sparse and discrete embeddings.
SNNs naturally project continuous, high-precision inputs into low-precision, event-driven representations. This observation sparks our curiosity about an interesting research question:

\begin{center}
\textit{Can we go beyond viewing spiking neurons merely as simple low-power units, and utilize spiking representations to build efficient tokenized Graph Transformers?}
\end{center}

\begin{figure*}[th]
    \centering
    \includegraphics[width=0.9\textwidth]{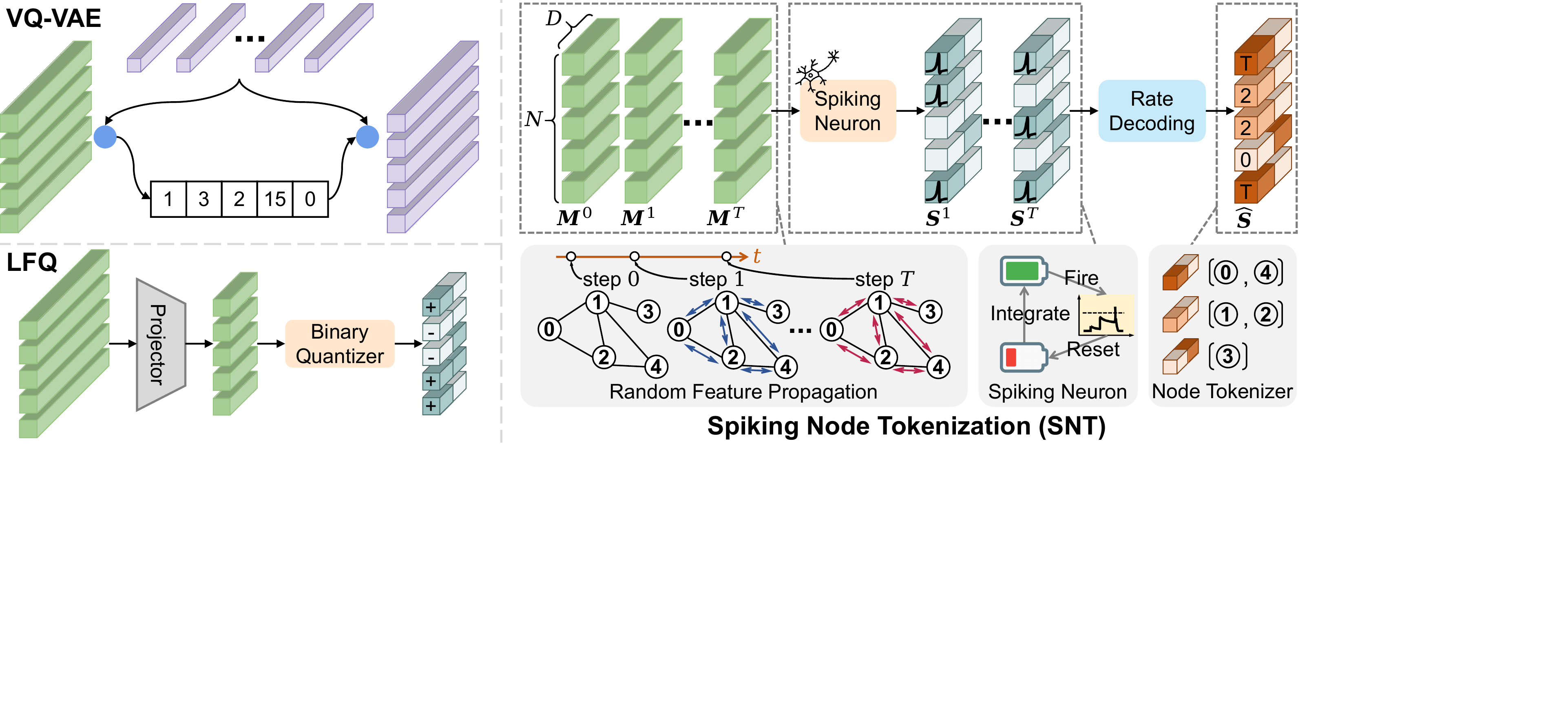}
    \caption{Frameworks of different tokenization strategies. Spiking Node Tokenization converts sequential inputs into spike count embeddings via spiking neurons, which yields a codebook-free, locality-aware node tokenization.}
\label{fig:1}
\end{figure*}

Tokenizers enable Transformer-based architectures to efficiently learn high-level representations from a small and manageable number of tokens. As shown in Figure~\ref{fig:1}, an exemplary implementation of tokenizers is VQ-VAE \cite{van2017neural}, which retrieves codewords from a fixed codebook to replace original embeddings. To alleviate the issue of under-utilization, Lookup Free Quantization (LFQ) \cite{yu2023language} proposes a learnable scalar quantization mechanism to build an implicit codebook. Inspired by existing tokenization strategies, we argue that \textbf{SNNs inherently convert sequential inputs into sparse and binary spike trains, which not only reduces energy consumption but also offers a novel sequence-to-token approach}. 
An energy-efficient node tokenization driven by SNNs is demonstrated in Figure~\ref{fig:1}. Spiking Node Tokenization (SNT) feeds full-precision embeddings from each propagation into spiking neurons. Spike count representations accumulated from spiking outputs represent nodes by a finite set of integer vectors, which we refer to as codewords.

Built upon the spike-driven node tokenizer, we propose GT-SNT, a linear-time \textbf{G}raph \textbf{T}ransformer based on \textbf{S}piking \textbf{N}ode \textbf{T}okenization for large-scale graphs. Specifically, GT-SNT balances local and global representations via two key components: Spiking Node Tokenization (SNT) and Codebook Guided Self-Attention (CGSA). SNT converts random feature propagation sequences into discrete, locality-aware node tokens. CGSA computes node-to-token attention scores based on generated spike count representations to capture long-range interactions in linear complexity.
GT-SNT features several advantages: a) Different from existing tokenized GTs, SNT proposes a generative and adaptive tokenization scheme built upon spiking neurons to evade the issue of \textit{codebook collapse}. b) SNT encodes graph inductive bias through well-designed sequential data. It leverages rich spiking dynamics and enables locality-aware quantization beyond simple binary quantization. c) CGSA is an efficient attention module by dynamically deriving the Key matrix from spike count representations without introducing complex machinery (e.g., distance metrics or auxiliary losses).
The contributions of this paper are summarized as follows:

\begin{itemize}
\item We propose Spiking Node Tokenization, which links SNNs with the sequential positional encoding data to generate locality-aware node tokens. To the best of our knowledge, our work is the first to explore the usage of SNN as an energy-saving tokenizer for graph data.
\item We tailor a node-to-token attention module with linear complexity, CGSA, which captures global topological information upon spike count node tokens. It provides a straightforward global aggregation guided by node tokens to balance both efficiency and expressiveness.
\item We conduct a comprehensive comparison with state-of-the-art baselines across graphs with various scales. Extensive experiments show that GT-SNT achieves comparable or even superior predictive performance on most datasets. Besides, GT-SNT enjoys up to \textbf{130x} faster inference speed compared to other GT baselines.
\end{itemize}

\section{Preliminaries}
\paragraph{Graph Neural Network.} We represent a graph as $\mathcal{G}=(\mathcal{V},\mathcal{E})$, where $\mathcal{V}$ is a set of nodes and $\mathcal{E}$ is a set of edges among these nodes, $\mathbf{A} \in \mathbb{R}^{N \times N}$ is the adjacency matrix of the graph. Let $N$ denote the number of nodes. We define the $d$-dimension nodes' attribute as $\mathbf{X} \in \mathbb{R}^{N \times d}$, which is known as the node feature matrix. For a given node $u\in \mathcal{V}$, GNN aggregates messages from its immediate neighborhood $N(u)$ and updates the node embedding $h_u$. This message passing process can be formulated as follows:
\begin{align}
    h^l_u&=\operatorname{UPDATE}(h^{l-1}_u, \operatorname{AGGREGATE}({h^l_v, \forall v\in\mathcal{N}(u)})),
\end{align}
where $h^l_u$ denotes the updated embedding of node $u$ in the $l$-th layer. $h^{l-1}_u$ is the embedding from the previous layer. $\operatorname{UPDATE}$ and $\operatorname{AGGREGATE}$ can be arbitrary differentiable functions.

\paragraph{Self-Attention.} As the most prominent component in the Transformer, the self-attention mechanism can be seen as mapping a query vector to a set of key-value vector pairs and calculating a weighted sum of value vectors as outputs. let nodes' attribute $\mathbf{X} \in\mathbb{R}^{N \times d}$ be the input to a self-attention layer. The attention function is defined as follows:
\begin{equation}\label{eq:7}
    \operatorname{Attn}(\mathbf{X})=\operatorname{softmax}(\frac{\mathbf{Q} \mathbf{K}^T}{\sqrt{d^{\prime}}})\mathbf{V},
\end{equation}
\begin{equation}\label{eq:8}
    \mathbf{Q}=\mathbf{X}\mathbf{W_q},\mathbf{K}=\mathbf{X}\mathbf{W_k},\mathbf{V}=\mathbf{X}\mathbf{W_v},
\end{equation}
where Query, Key and Value are calculated by learnable projection matrices $W_q, W_k, W_v \in\mathbb{R}^{d \times d^{\prime}}$. We omit the scaling factor hereafter for brevity.

\paragraph{Spiking Neural Network.} Although electrophysiological measurements can be accurately calculated by complex conductance-based neurons, the complexity limits their widespread deployment in deep neural networks. A simplified computational unit, which is known as the Integrate-and-Fire neuron, has been proposed \cite{salinas2002integrate}. IF neurons have three basic characteristics: \textbf{Integrate}, \textbf{Fire} and \textbf{Reset}. Firstly, the neuron integrates synaptic inputs from other neurons or external current $I$ to charge its cell membrane. Secondly, when the membrane potential reaches a pre-defined threshold value $V_{th}$, the neuron fires a spike $S$. Thirdly, the membrane potential of the neuron will be reset to $V_{reset}$ after firing. The neuronal dynamics can be formulated as follows:
\begin{align}\label{eq:1}
    \textbf{Integrate:} & \; V^t=\Psi(V^{t-1}, I^t)=V^{t-1} + I^t, \\
    \textbf{Fire:} & \; 
    S^t = \Theta(V^t-V_{th})=\left\{
    \begin{array}{ll}
         1,& V^t - V_{th} \geq 0 \\
         0,& \text{otherwise}
    \end{array},\right. \\
    \textbf{Reset:} & \; V^{t} = V^t(1 - S^t) + V_{reset} S^t,   
\end{align}
where $V^t$ and $I^t$ denote the membrane potential and input current at time step $t$, respectively. $\Theta(\cdot)$ denotes the fire function, implemented as a Heaviside step function. $\Psi(\cdot)$ is the membrane potential update function. Besides, there are two common IF variants, LIF and PLIF \cite{gerstner2014neuronal, fang2021incorporating}. The update function of these neurons can be formalized as follows:
\begin{align}\label{eq:2}
    \textbf{LIF:} & \; V^t = V^{t-1} + \frac{1}{\tau}(I^t-(V^{t-1}-V_{reset})), \\
    \textbf{PLIF:} & \; V^t = V^{t-1} + \frac{1}{1+\exp(-\beta)}(I^t-(V^{t-1}-V_{reset})),
\end{align}
where $\tau$ is the membrane time constant and $\beta$ is a trainable parameter. They are used to regulate how fast the membrane potential decays. The rate coding scheme is widely adopted by SNNs to extract spiking dynamics from sequential data. In this paper, we transform spiking outputs into spike counts that accumulate $S^t$ over $T$ time steps, $\hat{S}=\sum_{t=0}^TS^t$.
Besides, we adopt surrogate gradients during error backpropagation to address the issue of zero gradients caused by non-differentiable functions \cite{neftci2019surrogate}. The surrogate gradient method can be defined as $\Theta^{\prime}(x) \triangleq \theta^{\prime}(\alpha x)$, where $\alpha$ represents a smooth factor and $\theta(\cdot)$ represents a surrogate function \cite{neftci2019surrogate}.

\section{GT-SNT}

\begin{figure*}
    \centering
    \includegraphics[width=0.9\textwidth]{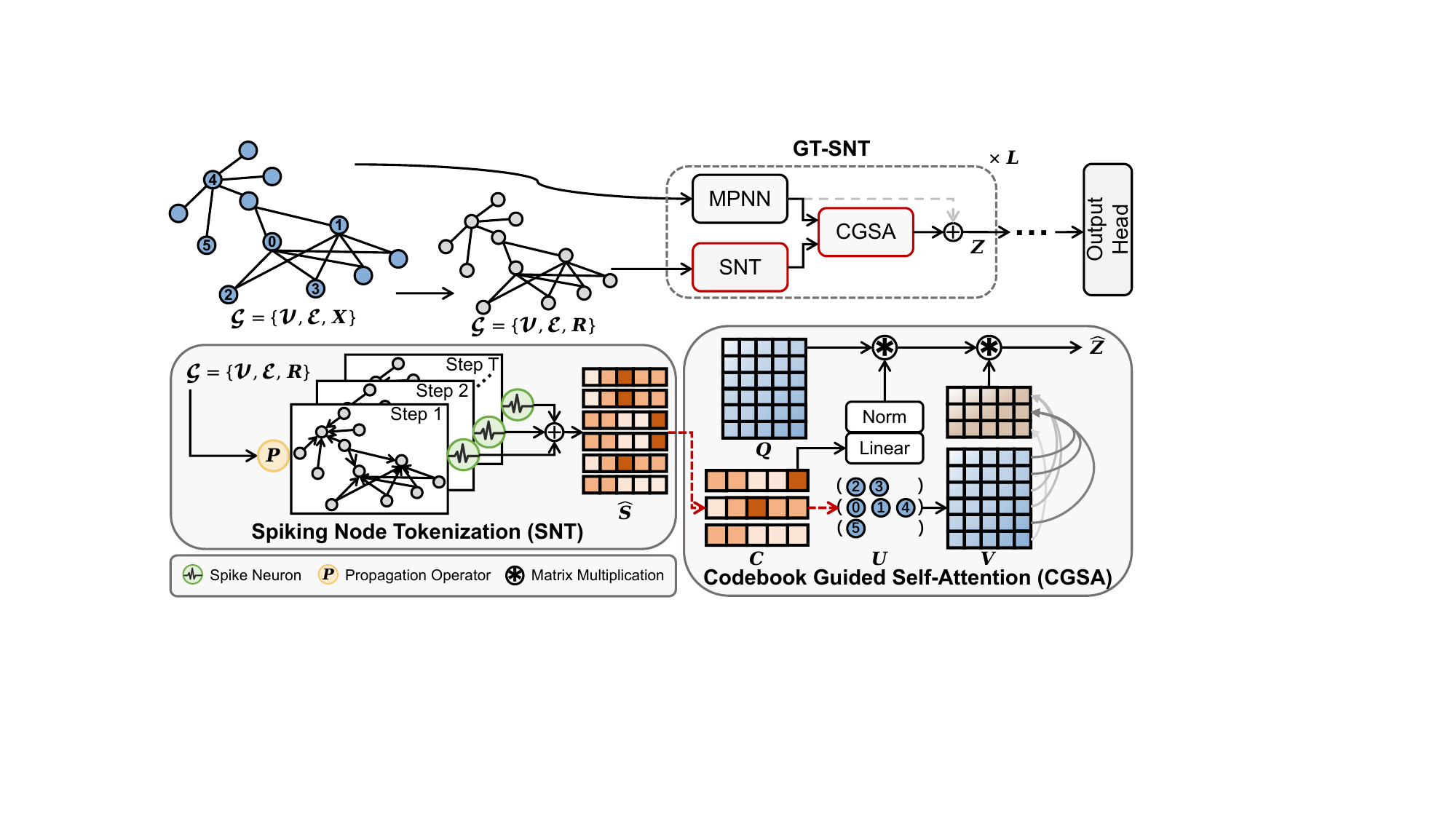}
    \caption{The overview of GT-SNT. Intuitively, the spiking node tokenization maps nodes to spike count vectors, which essentially groups nodes by their tokens. The codebook reconstructed from the spike count embeddings is fed into the codebook guided self-attention. It assists the self-attention block in calculating attention scores from nodes to grouped node sets in linear time and memory complexity.}
    \label{fig:2}
\end{figure*}

In this section, we comprehensively detail our approach, GT-SNT. As depicted in Figure~\ref{fig:2}, GT-SNT feeds the intermediate embeddings generated from feature propagation into the Spiking Node Tokenization (SNT) to inject graph positional information into the spike count embeddings. These codewords will guide the aggregation process in the self-attention module. Additionally, auxiliary message passing neural networks serve as encoders that encapsulate both node features and local graph topology to the attention module. In what follows, we detail the implementation of SNT. Then, we highlight how CGSA dynamically reconstructs a codebook only containing used codewords to apply node-to-token attention on large-scale graphs. Finally, we revisit the entire architecture of GT-SNT one by one.

\subsection{Spiking Node Tokenization}\label{subsec:4_1}
Typically, spiking neural networks are designed for time-varying data. In order to meet the requirement for time-varying data, a common practice in spiking graph neural networks is to repeatedly pass the static graph to spiking neurons at each time step. It inevitably brings high computational and storage overheads. In this work, we collect the node embeddings over $T$ propagation steps as the sequential inputs required for spiking neurons, which effectively reduce additional overheads. Besides, we decompose spike count embeddings accumulated from spiking outputs to dynamically reconstruct codebooks. This avoids the issue of explicitly defining the entire codebook that contains a large number of unused codewords.

Specifically, \textbf{(i)} we sample a $D$-dimension learnable random feature matrix $\mathbf{R} \in \mathbb{R}^{N \times D}$ from a uniform distribution, and define a propagation operator $\mathbf{P}$. Based on $\mathbf{P}$ and $\mathbf{R}$, we update and collect node embeddings $\mathbf{M}\in\mathbb{R}^{N \times D}$ over $T$ propagation steps, $\mathcal{M}=\{\mathbf{M}^0, \mathbf{M}^1,...,\mathbf{M}^T\}$. \textbf{(ii)} Given a SNN, it converts the above sequential node embeddings into spiking outputs $\mathcal{S}=\{\mathbf{S}^0, \mathbf{S}^1,...,\mathbf{S}^T\},\mathbf{S}\in\{0,1\}^{N \times D}$. \textbf{(iii)} The spike count embeddings $\hat{\mathbf{S}}\in\mathbb{Z}^{N \times D}$ are obtained by summing $\mathbf{S}^t$ over $T$ propagation steps. \textbf{(iv)} $\hat{\mathbf{S}}$ can be decomposed into the codebook $\mathbf{C}\in\mathbb{Z}^{B\times D}$ and the one-hot matrix $\mathbf{U}\in\{0,1\}^{N\times B}$. The above processes are defined as:
\begin{align}
    \mathbf{M}^t &= \operatorname{Norm}(\mathbf{P}\mathbf{M}^{t-1}), \quad\mathbf{M}^0 = \mathbf{R}, \\
    \mathbf{S}^t &= \Theta(\Psi(V^{t-1}, \mathbf{M}^t)-V_{th}), \\
    \hat{\mathbf{S}} &=\sum^T_{t=0}{\mathbf{S}^t},  \\
    \mathbf{U}\mathbf{C} &= \hat{\mathbf{S}},
\end{align}
where $Norm(\cdot)$ normalizes output messages to the range of threshold membrane potential. $\Theta(\cdot)$ and $\Phi(\cdot)$ are membrane potential update and fire functions, respectively. We choose similar propagation operators presented in previous work \cite{eliasof2023graph}. The codebook $\mathbf{C}$ is created by removing duplicate row vectors in $\hat{\mathbf{S}}$. The one-hot matrix $\mathbf{U}$ can be seen as the indices for where the codewords in $\hat{\mathbf{S}}$ ended up in $\mathbf{C}$. Hereafter, we will refer to $B$ as the size of the reconstructed codebook.

We define a discrete latent space $\tilde{\mathcal{C}}$ of the spike count vector $\hat{s}\in\mathbb{Z}^D$ , $\forall \hat{s}\in\tilde{\mathcal{C}}$. The size of $\tilde{\mathcal{C}}$ is determined by both the number of propagation steps $T$ and the dimensionality of each random feature $D$, $|\tilde{\mathcal{C}}|=(T+1)^D$. This is because each dimension can take $T+1$ possible spike counts (considering the case where the total number of spikes is zero) across $T$ time steps. Different from vanilla VQ methods, which explicitly pre-define a codebook containing all latent embedding vectors for lookup, SNT dynamically maps continuous full-precision embedding to the subset of $\tilde{\mathcal{C}}$ via the SNN. It enables the size of the reconstructed codebook to be much smaller, $B\ll |\tilde{\mathcal{C}}|$, while effectively alleviating the \textit{codebook collapse} problem.

\subsection{Codebook Guided Self-Attention}\label{subsec:4_2}
After the Spiking Node Tokenization, we propose a codebook guided self-attention (CGSA) to capture long-range signals in linear complexity. Specifically, node embeddings $\mathbf{H}\in\mathbb{R}^{N \times d^{\prime}}$ from auxiliary MPNNs are introduced as Query and Value. We materialize the Key $\hat{\mathbf{K}}$ based on the matrices $\mathbf{C}$ and $\mathbf{U}$. The attention functions are defined as:
\begin{align}
    \mathbf{Q}&=\mathbf{H}\mathbf{W_q}, \quad\mathbf{V}=\mathbf{H}\mathbf{W_v}, \\
    \mathbf{G} &=  \operatorname{Norm}(\operatorname{Linear}(\mathbf{C})), \\
    \hat{\mathbf{K}} &= \mathbf{U}\mathbf{G}, \\
    \hat{\mathbf{Z}} &= \operatorname{softmax}(\mathbf{Q}\hat{\mathbf{K}}^T)\mathbf{V}, \label{eq:16}
\end{align}
where $d^{\prime}$ denotes the dimension of intermediate embeddings. In CGSA, $\hat{\mathbf{K}}$ is calculated based on the matrix multiplication between $\mathbf{G}$ and $\mathbf{U}$, rather than performing the nearest neighbor look-up on the entire codebook. It avoids the severe reliance on complex components like distance measures or auxiliary losses. Derived from \cite{lingle2023transformer}, the attention weights in Eq~\ref{eq:16} can be further factored:
\begin{align}
\hat{\mathbf{Z}} &= \operatorname{softmax}(\mathbf{Q}\hat{\mathbf{K}}^T)\mathbf{V} \nonumber\\
&=\operatorname{softmax}(\mathbf{Q}(\mathbf{U}\mathbf{G})^T)\mathbf{V} \nonumber\\
&=\operatorname{Diag}^{-1}(\operatorname{exp}(\mathbf{Q}\mathbf{C}^T)\mathbf{U}^T\mathbf{1})\operatorname{exp}(\mathbf{Q}\mathbf{C}^T) \mathbf{U}^T \mathbf{V}
\end{align}
where $\mathbf{1}\in\mathbb{R}^N$. $\mathbf{U}^T\mathbf{1}=\{n_b\}^B_b\in\mathbb{Z}^+$ denotes the number of spike count embeddings in $\hat{\mathbf{S}}$ mapped to the same codewords. The complexity of CGSA is $O(NBd_{v})$, where $B \ll N$. It can be considered that the computational overhead of CGSA grows linearly with the number of nodes. To avoid generating an excessively large codebook in the initial phase of learning, we perform a truncation strategy. We rank ${n_b}$ from high to low and select the top $B_{max}$ to generate a truncated codebook, which ensures the training efficiency of GT-SNT.

\subsection{Overall Framework}\label{subsec:4_3}
As shown in Figure~\ref{fig:2}, the overview of GT-SNT includes four modules: SNT, auxiliary MPNN, CGSA and a classification head (CH). In the SNT, we construct learnable random features and spiking neurons for each layer. By defining a shared propagation operator, messages among nodes are collected and transformed into spike count embeddings $\hat{\mathbf{S}}$. Then auxiliary MPNNs as encoders generate node embeddings with semantic and local neighborhood information. In the CGSA, the reconstructed codebook $\mathbf{C}$, codeword indices $\mathbf{U}$ and node embeddings $\mathbf{H}$ are fed into a linear-time self-attention to capture global topological information. Different from the vanilla Transformer, CGSA employs global node-to-token attention to actively introduce graph inductive bias. These four parts can be written as follows:
\begin{align}
    \mathbf{U}^l,\mathbf{C}^l &= \operatorname{SNT}^l(\mathbf{A}), \\
    \mathbf{H}^l &= \operatorname{MPNN}^l(\mathbf{Z}^{l-1}, \mathbf{A}), \\
    \hat{\mathbf{Z}}^l &= \operatorname{CGSA}^l(\mathbf{U}^l, \mathbf{C}^l, \mathbf{H}^l), \\
    \mathbf{Z}^l &= \operatorname{Linear}(\hat{\mathbf{Z}}^l) + \mathbf{H}^l, \\
    \mathbf{Y} &= \operatorname{CH}(\mathbf{Z}^L),
\end{align}
where $L$ is the number of layers. We choose a simple fully connected layer as the classification head. The auxiliary MPNNs are implemented as a single-layer GCN without normalization. It has been observed that projection blocks, including Multilayer Perceptrons (MLPs) and normalization layers, exacerbate the overfitting problem on large-scale graphs for vanilla Transformers. Therefore, we discard projection layers and retain only the self-attention module and the residual connection \cite{he2016deep}.

\section{Experiments}
\subsection{Comparison with Existing Models}
In this section, we conduct the experimental evaluation to show the effectiveness of GT-SNT on nine node classification datasets, including three citation networks \cite{kipf2016semi}, two co-purchase networks \cite{shchur2018pitfalls}, two heterophilic graphs \cite{wu2024sgformer} and two OGB graphs \cite{hu2020open}. A head-to-head comparison is conducted with state-of-the-art GNNs and GTs, based on their architectural designs. As shown in Table~\ref{tab:1}, these baselines fall into three categories: spike-based methods, Graph Transformer framework and vector quantization-based methods. The hyperparameter search strategy is deployed on both GT-SNT and other baselines to get the optimal combinations of parameters. We perform all models on each dataset 5 times with different random seeds to report the mean and standard deviation of metrics. All experiments are conducted on a single NVIDIA RTX 4090 GPU. 

\paragraph{Overall performance.} The experimental results are demonstrated in Table~\ref{tab:2}. As shown in the table, our method achieves promising performance on all datasets. This is a significant advancement considering the information loss caused by low-precision spike count embeddings. Specifically, GT-SNT achieves predictive performances on par or even better than high-precision GT baselines. \textbf{In general, our method shows slight improvements on 5 out of 9 datasets}. GT-SNT effectively injects the graph inductive bias into the self-attention block while capturing long-range interactions, resulting in strong expressive power. Besides, we observe that GT-SNT has competitive performances on two heterophily datasets, particularly Actor. It suggests that GT-SNT can also maintain strong expressive power on heterophilic graphs. \textbf{GT-SNT outperforms other spike-based baselines on all datasets, which achieves an average improvement of 3.9\%.} Constructing the input sequence from propagation steps rather than repeating graphs multiple times endows GT-SNT with desirable scalability. In addition, GT-SNT achieves a better trade-off between latent space compression and information retention by introducing spiking neurons as the node tokenizer instead of artificial neuron substitutes.

\begin{table}[tbp]
    \centering
    \begin{tabular}{|c|ccc|}
        \hline
	\multirow{2}{*}{\textbf{Models}} & \multicolumn{3}{c|}{\textbf{Components}} \\
        \cline{2-4}
        & \textbf{SP} & \textbf{GT} & \textbf{VQ} \\
        \hline
        SpikingGCN\cite{zhu2022spiking} & \checkmark & - & - \\
        SpikeNet\cite{li2023scaling} & \checkmark & - & - \\
        SpikeGCL\cite{li2023graph} & \checkmark & - & - \\
        SpikeGT\cite{sun2024spikegraphormer} & \checkmark & \checkmark & - \\
        \hline
        NAGphormer\cite{chen2022nagphormer} & - & \checkmark & - \\
        NodeFormer\cite{wu2022nodeformer} & - & \checkmark & - \\
        SGFormer\cite{wu2024sgformer} & - & \checkmark & - \\
        GOAT\cite{kong2023goat}     & - & \checkmark & \checkmark \\
        VQGraph\cite{yang2024vqgraph}  & - & - & \checkmark \\
        \hline
        \textbf{GT-SNT} & \checkmark & \checkmark & \checkmark \\
        \hline
    \end{tabular}
    \caption{\label{tab:1} Comparison of Graph Transformers and Graph Neural Networks w.r.t. required components (\textbf{SP}: spike-based, \textbf{GT}: Graph Transformer framework, \textbf{VQ}: vector quantization-based).}
\end{table}

\begin{table*}
    \centering
    \tabcolsep=5pt
    \begin{tabular}{lccccccccccc}
        \toprule
        \textbf{Models}  & \textbf{Cora} & \textbf{CiteSeer} & \textbf{PubMed} & \textbf{Co-CS} & \textbf{Co-Physics} & \textbf{Actor} & \textbf{Deezer} & \textbf{arXiv} & \textbf{Products} \\
        $\#nodes$ & 2,708 & 3,327 & 19,717 & 18,333  & 34,493 & 7,600 & 28,281 & 169,343 & 2,449,029 \\
        $\#edges$ & 10,556 & 9,104 & 88,648 & 163,788 & 495,924 & 30,019 & 185,504 & 1,166,243 & 61,859,140 \\
        \midrule
        GCN     & 81.6$_{\pm0.4}$ & 71.6$_{\pm0.4}$ & 78.8$_{\pm0.6}$ & 92.5$_{\pm0.4}$ & 95.7$_{\pm0.5}$ & 30.1$_{\pm0.2}$ & 62.7$_{\pm0.7}$ & 70.4$_{\pm0.3}$ & 75.7$_{\pm0.1}$ \\
        GAT     & 83.0$_{\pm0.7}$ & 72.1$_{\pm1.1}$ & 79.0$_{\pm0.4}$ & 92.3$_{\pm0.2}$ & 95.4$_{\pm0.3}$ & 29.8$_{\pm0.6}$ & 61.7$_{\pm0.8}$ & 70.6$_{\pm0.3}$ & OOM \\
        SGC     & 80.1$_{\pm0.2}$ & 71.9$_{\pm0.1}$ & 78.7$_{\pm0.1}$ & 90.3$_{\pm0.9}$ & 93.2$_{\pm0.5}$ & 27.0$_{\pm0.9}$ & 62.3$_{\pm0.4}$ & 68.7$_{\pm0.1}$ & 74.2$_{\pm0.1}$ \\
        VQGraph & 81.1$_{\pm1.2}$ & \first{74.5$_{\pm1.9}$} & 77.1$_{\pm3.0}$ & 93.3$_{\pm0.1}$ & 95.0$_{\pm0.1}$ & \second{38.7$_{\pm1.6}$} & 65.1$_{\pm0.2}$ & \second{72.4$_{\pm0.2}$} & \second{78.3$_{\pm0.1}$} \\
        \midrule
        SpikingGCN & 79.1$_{\pm0.5}$ & 62.9$_{\pm0.1}$ & 78.6$_{\pm0.4}$ & 92.6$_{\pm0.3}$ & 94.3$_{\pm0.1}$ & 26.8$_{\pm0.1}$ & 58.2$_{\pm0.3}$ & 55.8$_{\pm0.7}$ & OOM \\
        SpikeNet   & 78.4$_{\pm0.7}$ & 64.3$_{\pm0.8}$ & 79.1$_{\pm0.5}$ & 93.0$_{\pm0.1}$ & 95.8$_{\pm0.7}$ & 36.2$_{\pm0.9}$ & 65.0$_{\pm0.2}$ & 66.8$_{\pm0.1}$ & 74.3$_{\pm0.4}$ \\
        SpikeGCL   & 79.8$_{\pm0.7}$ & 64.9$_{\pm0.2}$ & 79.4$_{\pm0.8}$ & 92.8$_{\pm0.1}$ & 95.2$_{\pm0.6}$ & 30.3$_{\pm0.5}$ & 65.0$_{\pm1.1}$ & 70.9$_{\pm0.1}$ & OOM \\
        SpikeGT & 82.0$_{\pm0.7}$ & 70.5$_{\pm0.6}$ & 71.1$_{\pm0.4}$ & 92.1$_{\pm0.8}$ & 95.7$_{\pm0.3}$ & 36.0$_{\pm0.5}$ & 65.6$_{\pm0.2}$ & 70.2$_{\pm0.9}$ & OOM \\
        \midrule
        NAGphormer & 79.9$_{\pm0.1}$ & 68.8$_{\pm0.2}$ & \second{80.3$_{\pm0.9}$} & 93.1$_{\pm0.5}$ & 95.7$_{\pm0.7}$ & 33.0$_{\pm0.9}$ & 64.4$_{\pm0.6}$ & 70.4$_{\pm0.3}$ & 73.3$_{\pm0.7}$ \\
        GOAT & 78.6$_{\pm0.5}$ & 68.4$_{\pm0.7}$ & 78.1$_{\pm0.5}$ & \second{93.5$_{\pm0.6}$} & 95.4$_{\pm0.2}$ & 37.5$_{\pm0.7}$ & 65.1$_{\pm0.3}$ & \second{72.4$_{\pm0.4}$} & \first{82.0$_{\pm0.4}$} \\
        NodeFormer  & 82.2$_{\pm0.9}$ & 72.5$_{\pm1.1}$ & 79.9$_{\pm1.0}$ & 92.9$_{\pm0.1}$ & 95.4$_{\pm0.1}$ & 36.9$_{\pm1.0}$ & \second{66.4$_{\pm0.7}$} & 59.9$_{\pm0.4}$ & 72.9$_{\pm0.1}$ \\
        SGFormer  & \second{84.5$_{\pm0.8}$} & 72.6$_{\pm0.2}$ & \second{80.3$_{\pm0.6}$} & 91.8$_{\pm0.2}$ & \second{95.9$_{\pm0.8}$} & 37.9$_{\pm1.1}$ & \first{67.1$_{\pm1.1}$} & \first{72.6$_{\pm0.1}$} & 72.6$_{\pm1.2}$ \\
        \midrule
        GT-SNT & \first{84.7$_{\pm0.8}$} & \second{74.0$_{\pm0.5}$} & \first{80.6$_{\pm0.4}$} & \first{93.7$_{\pm0.4}$} & \first{96.2$_{\pm0.0}$} & \first{39.1$_{\pm0.2}$} & 65.7$_{\pm0.1}$ & \second{72.4$_{\pm0.3}$} & 74.8$_{\pm0.4}$\\
        \bottomrule
    \end{tabular}
    \caption{Classification accuracy(\%) on nine datasets. Highlighted are the top \first{first}, \second{second} results.}\label{tab:2}
\end{table*}

\subsection{Characteristics of Spiking Node Tokenization}\label{sec:4_2}
To thoroughly analyze the spiking node tokenization, we conduct a series of experiments on the GT-SNT. We track the metrics including Codebook Usage and Accuracy to explore the following questions: \textbf{(i)} How does the size of the latent embedding space affect GT-SNT? \textbf{(ii)} Is the spiking node tokenization a more efficient tokenization alternative?

\begin{figure}[t]
\centering
    \centering
    \includegraphics[width=\linewidth]{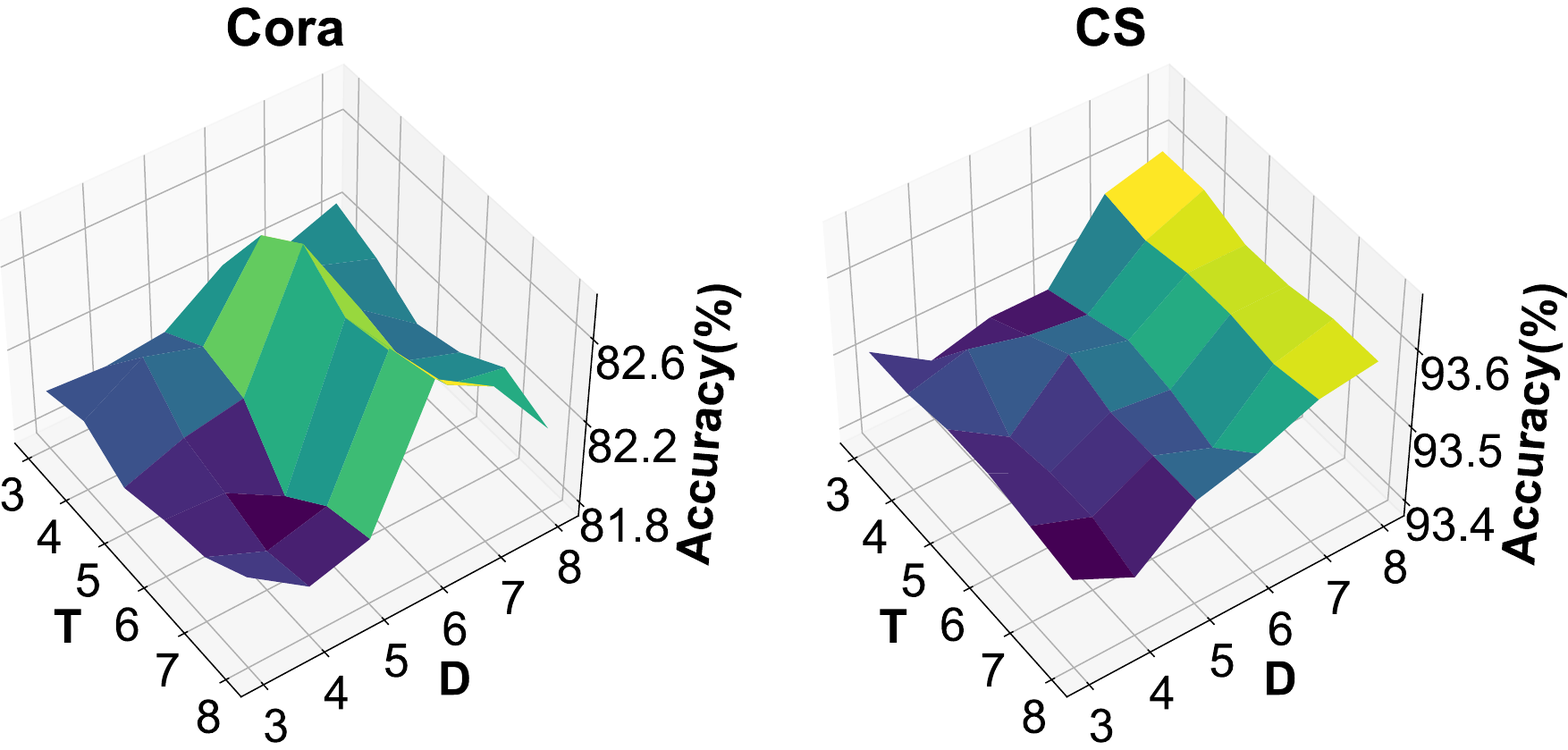}
    \caption{Influences of random feature dimensionality $D$ and propagation step $T$ on Cora and CS.}
    \label{fig:3}
\end{figure}

\paragraph{The influence of $|\tilde{\mathcal{C}}|$.} As aforementioned above, the number of propagation steps $T$ and the random feature dimension $D$ determine the size of the latent space. In Figure~\ref{fig:3}, we evaluate the performances of GT-SNT under different combinations between $D$ and $T$. The visualization results show that an excessively small latent space size may impair the performance of GT-SNT. It generally improves accuracy as the latent space size increases. Essentially, $|\tilde{\mathcal{C}}|$ determines the upper bound of information the reconstructed codebook can store. When $|\tilde{\mathcal{C}}|$ is much smaller than the number of nodes, indiscriminately mapping a large number of nodes to the same codeword will bring significant information loss. Besides, we observe the diminishing gains from latent space size scaling, especially on small-scale graphs. We believe that node embeddings merely are transformed into low-precision counterparts rather than being compressed appropriately when the feature dimensionality $D$ is too large.
The propagation step $T$ correlates with the complexity of spiking patterns. As depicted in Figure~\ref{fig:3}, it suggests that $T > 6$ may lead to subpar performance. Notably, GT-SNT outperforms most spike-based baselines even under the worst parameter combinations. It demonstrates that GT-SNT evades the difficulty of codebook learning and provides a robust, easy-to-use node tokenization approach. 

\begin{figure}[t]
    \centering
    \includegraphics[width=\linewidth]{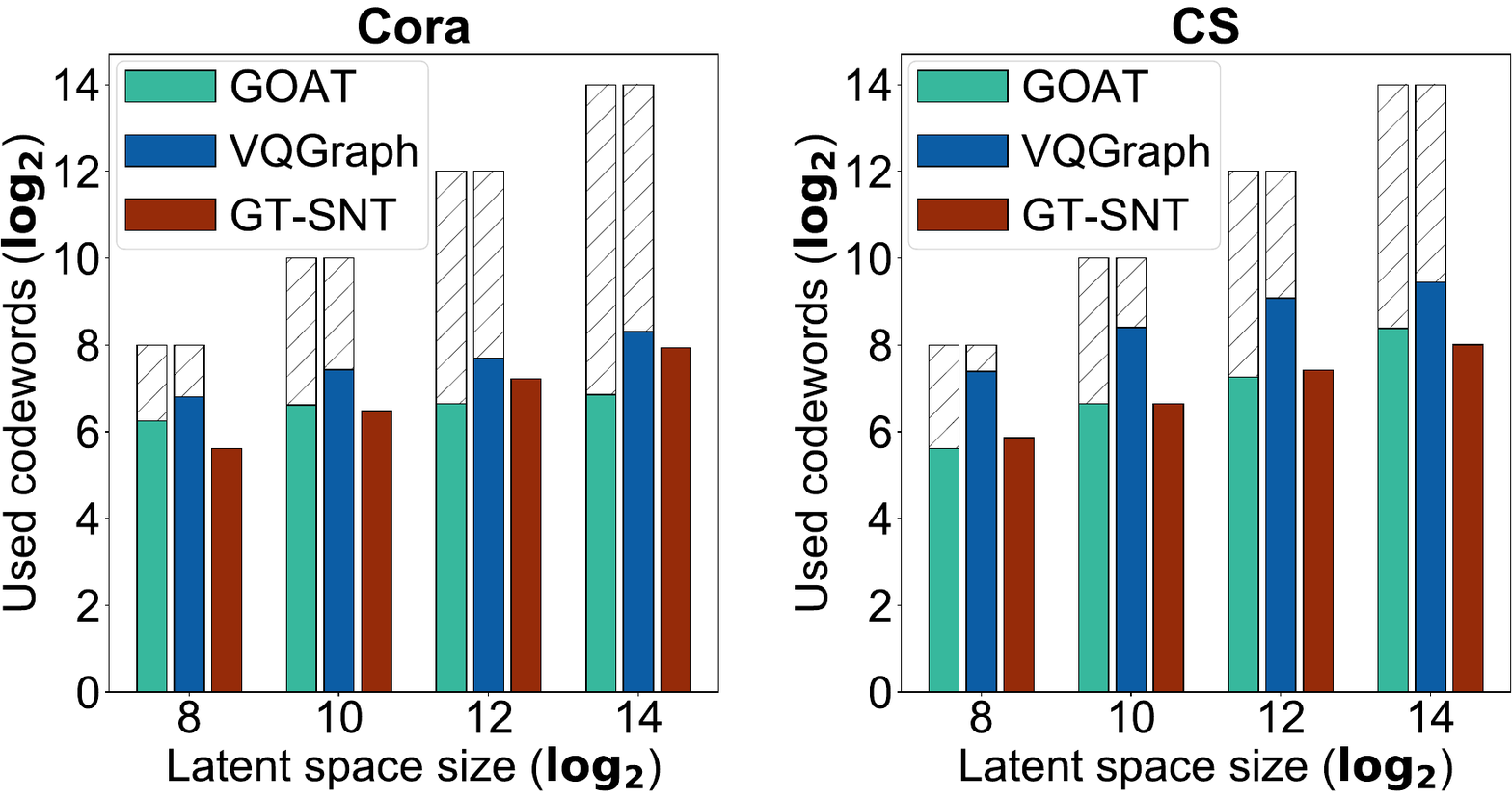}
    \caption{Codebook usage under four different latent space sizes on Cora and CS datasets. The solid-colored bar represents the number of used codewords, while the hatched bar denotes the size of the pre-defined codebook.}
    \label{fig:4}
\end{figure}

\paragraph{Codebook Usage.} 
For exploring the \textit{codebook collapse} problem in VQ-based graph models, we track the fraction of codewords that are used at least once after the training phase. The codebook usages between GT-SNT and the other VQ-based graph methods are shown in Figure~\ref{fig:4}. In previous works, the pre-defined codebook is used to represent the entire discrete latent space. We select four combinations of $D$ and $T$ ($T=3/D=4$, $T=3/D=5$, $T=3/D=6$, $T=3/D=7$) to match the latent space sizes ($2^{8}$, $2^{10}$, $2^{12}$, $2^{14}$).
The results demonstrate that those methods using pre-defined codebooks suffer from the serious issue of \textit{codebook collapse}. As the pre-defined codebook size increases, the codebook usage decreases. For GOAT, the average codebook usages are 10.6\% and 8.7\% on Cora and CS datasets. The codebook usages in VQGraph are slightly higher, which achieve 16.9\% and 29.0\%. When the latent space size exceeds $2^{10}$, the usages of both methods drop below 50\%. It reveals that these VQ-based graph learning methods struggle with efficiently using large pre-defined codebooks. 
SNT provides a generative solution, which dynamically selects the subset from a discrete spike count embedding space to reconstruct the codebook. It brings 100\% codebook usage. Although the number of used codewords in GT-SNT is slightly larger than that of vanilla VQ counterparts in some cases, it is still significantly smaller than the pre-defined codebooks. In a nutshell, SNT achieves better codebook utilization by viewing spiking neurons as a sequence-to-token approach.

\begin{table*}[!ht]
    \centering
    \resizebox{\linewidth}{!}{
        \begin{threeparttable}
        \begin{tabular}{l|c|ccccc|cc}
            \toprule
            \textbf{Datasets} & \textbf{Metrics} & \textbf{NAGphormer} & \textbf{GOAT} & \textbf{NodeFormer} & \textbf{SGFormer} & \textbf{SpikeGT} & \textbf{Avg} & \textbf{GT-SNT} \\
            \midrule
            \multirow{3}{*}{CS} & Latency$\downarrow$ & 0.70 & 5.02 & 0.05 & \first{0.01} & \second{0.03} & 1.16 & \first{0.01} (\underline{$\times$116}) \\
            & Memory$\downarrow$ & 3400 & 12490 & 2822 & \second{1662} & 8542 & 5783 & \first{1638} \\
            & Energy$\downarrow$ & 0.82 & 1.21 & \second{0.21} & 0.35 & 0.59 & 0.64 & \first{0.16} \\
            \midrule
            \multirow{3}{*}{Physics} & Latency$\downarrow$ & 1.79 & 10.98 & 0.14 & \first{0.02} & \second{0.08} & 2.60 & \first{0.02} (\underline{$\times$130}) \\
            & Memory$\downarrow$ & 13628 & 22776 & 7624 & \first{2944} & 16414 & 12677 & \second{3036} \\
            & Energy$\downarrow$ & 1.86 & 2.35 & \second{0.46} & 0.78 & 1.35 & 1.36 & \first{0.37} \\
            \midrule
            \multirow{3}{*}{arXiv} & Latency$\downarrow$ & 0.78 & 28.27 & 1.17 & \second{0.10} & 0.30 & 6.12 & \first{0.08} (\underline{$\times$77}) \\
            & Memory$\downarrow$ & 10450 & 21146 & 11988 & \first{6386} & 22654 & 14525 & \second{7132} \\
            & Energy$\downarrow$ & 1.12 & 9.92 & 0.63 & 0.57 & \first{0.13} & 2.47 & \second{0.18} \\
            \midrule
            \multirow{3}{*}{Products} & Latency$\downarrow$ & 25.74 & 2416.84 & 41.78 & \second{24.34} & - & 627.18 & \first{20.83} (\underline{$\times$30}) \\
            & Memory$\downarrow$ & \second{7470} & 21974 & 10500 & \first{934} & - & 10220 & 13494 \\
            & Energy$\downarrow$ & 16.06 & 143.80 & 10.05 & \second{8.07} & - & 44.50 & \first{3.69} \\
            \bottomrule
        \end{tabular}
        \end{threeparttable}
    }
    \caption{The maximum memory usage (MB), theoretical energy consumption (J) and inference latency (s) of various GT methods. The speedup over average inference latency is underlined.}\label{tab:3}
\end{table*}

\subsection{Energy Efficiency Analysis}
To verify the efficiency of GT-SNT, we conduct energy efficiency analysis using the following metrics: peak memory usage during training, inference latency and theoretical energy consumption. The theoretical energy consumption estimation is derived from \cite{yao2024spike}. The results are presented in Table~\ref{tab:3}.

The results show that GT-SNT achieves the lowest inference latency across all datasets. \textbf{Compared to other GTs, GT-SNT with better performance achieves up to 130x lower inference latency.}
Most VQ-based methods tend to retrieve and replace node representations with the closest codewords during the inference phase. In SNT, trained spiking neurons directly convert input features into codewords, which reduces the computational cost brought by retrieval. Integrating the reconstructed codebook in SNT with linear-time global attention brings a significant improvement in inference speed.
For large-scale graphs with high feature dimensionality, repeating the graph to generate sequential inputs leads to high energy consumption. The results show that previous spike-based GT is hard to manifest its superiority in energy consumption on CS and Physics. Through generating low-dimensional random features and collecting sequential embeddings from propagation steps, \textbf{GT-SNT outperforms another spike-based GT in memory usage and latency across all datasets while maintaining acceptable theoretical energy consumption.}

\subsection{Ablation Study}
In this section, we conduct ablation studies to explore the influences of different components on predictive performances. We construct 8 baselines by combining two classic linear-time attention modules (Performer \cite{choromanski2020rethinking} and Linformer \cite{wang2020linformer}), two spiking neurons (IF and LIF), two normalization algorithms (LayerNorm \cite{ba2016layernormalization} and STFNorm \cite{xu2021exploiting}) and two VQ modules (VQ-VAE \cite{van2017neural} and FSQ \cite{mentzer2023finite}).

The experimental results are demonstrated in Table~\ref{tab:4}. Although incorporating extra positional encodings enables Performer and Linformer to handle graph prediction tasks, they struggle to achieve good predictive performance on large-scale graphs like ogbn-arxiv. In GT-SNT, the CGSA actively introduces the global positional encoding during attention score calculation. It suggests that developing topology-aware tokenized Transformers is a promising direction for scaling GTs on large-scale graphs. 
The choice of spiking neurons also affects the predictive performances of GT-SNT. PLIF models with learnable membrane time constants and synaptic weights achieve better performances in most cases. These neurons effectively improve the flexibility of SNT. In addition, the well-designed normalization algorithm for spiking neurons, STFNorm outperforms the LayerNorm algorithm across all datasets. 
For complex tasks, GT-SNT with dynamic learning and adaptive codebook granularity surpasses the simple quantizers such as FSQ. Benefitting from diverse spiking dynamics and finer quantization, GT-SNT outperforms both FSQ and full-precision VQ-VAE.

\begin{table}[thb]\small
    \centering
    \tabcolsep=5pt
    \begin{tabular}{l|cccc}
        \toprule
        \textbf{Models} & \textbf{PubMed} & \textbf{CS} & \textbf{Physics} & \textbf{ogbn-arxiv} \\
        \midrule
        +Performer & 80.2$_{\pm0.2}$ & 93.1$_{\pm0.4}$ & 95.8$_{\pm0.1}$ & 71.2$_{\pm0.1}$ \\
        +Linformer & 79.6$_{\pm1.0}$ & 92.6$_{\pm0.5}$ & 95.5$_{\pm0.1}$ & 65.2$_{\pm1.3}$ \\
        \midrule
        +IF  & 81.6$_{\pm1.2}$ & 92.8$_{\pm0.1}$ & 96.0$_{\pm0.4}$ & 71.0$_{\pm0.5}$ \\
        +LIF & 79.6$_{\pm0.7}$ & 92.8$_{\pm0.0}$ & 96.1$_{\pm0.2}$ & 72.1$_{\pm0.2}$ \\
        \midrule
        +LayerNorm & 78.9$_{\pm1.3}$ & 90.3$_{\pm0.6}$ & 95.4$_{\pm0.4}$ & 71.2$_{\pm0.2}$ \\
        +STFNorm   & \first{82.6$_{\pm0.2}$} & 92.3$_{\pm0.4}$ & \first{96.5$_{\pm0.5}$} & \first{72.4$_{\pm0.7}$} \\
        \midrule
        +VQ       & 79.8$_{\pm0.4}$ & 93.2$_{\pm0.3}$ & 95.0$_{\pm0.4}$ & 70.7$_{\pm0.7}$ \\
        +FSQ   & 78.4$_{\pm0.9}$ & \first{93.7$_{\pm0.2}$} & 95.8$_{\pm0.1}$ & 71.6$_{\pm0.2}$ \\
        \midrule
        GT-SNT & 80.6$_{\pm0.5}$ & \first{93.7$_{\pm0.4}$} & 96.2$_{\pm0.0}$ & \first{72.4$_{\pm0.3}$} \\
        \bottomrule
    \end{tabular}
    \caption{Ablation studies on PubMed, CS, Physics and ogbn-arxiv datasets. And $+x$ means replacing the original component in GT-SNT with $x$.}\label{tab:4}
\end{table}

\section{Conclusion}
In this study, we propose GT-SNT, a linear-time Graph Transformer via spiking node tokenization. We find that positional encoding patterns of different nodes can be encoded into the same codewords. Inspired by the observation, GT-SNT not only considers spiking neurons as the low-power units, but also incorporates SNNs as a learnable tokenizer into Graph Transformers. It enables GT-SNT to achieve faster inference speed and better predictive performance. Spiking node tokenization that dynamically creates a spike count-based codebook paves a different way to address issues within current VQ-based graph models. We believe that our work holds great promise for the development of brain-inspired neural networks, which bridge the gap between existing SNNs and node tokenization. We hope it will inspire further research into energy-saving Graph Transformers.

\section{Acknowledgments}
This work was supported in part by the National Key R\&D Program of China under Grant 2022YFF0902500, in part by the Guangdong Basic and Applied Basic Research Foundation, China under Grant 2023A1515011050, in part by Shenzhen Science and Technology Program under Grant KJZD20231023094501003, and in part by Tencent AI Lab under Grant RBFR2024004.

\bibliography{aaai2026}

\end{document}